%% file: main.tex
\documentclass[runningheads]{llncs}

\usepackage{eccv}

\usepackage{eccvabbrv}

\usepackage{graphicx}
\usepackage{booktabs}
\usepackage[inline,shortlabels]{enumitem}

\usepackage[accsupp]{axessibility}  %

\input{preamble}

\usepackage{hyperref}

\usepackage{orcidlink}

\begin{document}
\raggedbottom

\title{FullCircle: Effortless 3D Reconstruction \\ from Casual 360° Captures}

\titlerunning{FullCircle}

\author{
Yalda Foroutan$^\ast$\inst{1}   \and
Ipek Oztas$^\ast$    \inst{1, 2}\and
Daniel Rebain        \inst{3}   \and
Aysegul Dundar       \inst{2}   \and\\
Kwang Moo Yi         \inst{3}   \and
Lily Goli            \inst{4}   \and
Andrea Tagliasacchi  \inst{1, 4}
}

\authorrunning{Y.~Foroutan, I.~Oztas et al.}

\institute{
Simon Fraser University \and
Bilkent University \and
University of British Columbia \and
University of Toronto
}

\maketitle
\begingroup
\renewcommand\thefootnote{}
\footnotetext{$^\ast$ Equal contribution.}
\endgroup
\input{sec/0_abstract}

\input{sec/1_intro}
\input{sec/2_related}

\input{sec/3_method}
\input{sec/4_results}

\input{sec/5_conclusion}

\section*{Acknowledgements}
This work was supported by the Natural Sciences and Engineering Research Council of Canada (NSERC) Discovery Grant [2023-05617], the Canada Foundation for Innovation (CFI), John R. Evans Leaders Fund (JELF), the SFU Visual Computing Research Chair, and Digital Research Alliance of Canada.

{
    \small
    \bibliographystyle{splncs04}
    \bibliography{main}
}

\input{sec/X_suppl}

\end{document}

%% file: preamble.tex
\usepackage{cuted} %

\usepackage{currfile} %

\usepackage{caption} %
\usepackage{comment}

\definecolor{dark_green}{rgb}{0, 0.4, 0}

\newcommand{\at}[1]{{\color{black}#1}}

\newcommand{\yf}[1]{{\color{black}#1}}

\newcommand{\dr}[1]{{\color{black}#1}}

\newcommand{\lily}[1]{{\color{black}#1}}

\usepackage{soul}
\setuldepth{foobar}

\usepackage{pifont}
\newcommand{\cmark}{\ding{51}} %
\newcommand{\xmark}{\ding{55}} %

\definecolor{mypink}{RGB}{168,130,155} %

\renewcommand{\paragraph}[1]{\vspace{.25em}\noindent\textbf{#1}}

\newcommand{\methodname}{FullCircle\xspace}

\usepackage{multirow}
\usepackage{fontawesome5} %
\usepackage{overpic}
\definecolor{darkpurple}{RGB}{120, 70, 150}  %
\definecolor{darkyellow}{RGB}{220, 170, 0}   %

\usepackage{lipsum}
\usepackage{makecell}

\definecolor{lightred}{RGB}{235,70,85}      
\definecolor{darkpurple}{RGB}{140,90,200}

\usepackage{tikz}
\usepackage{xcolor}

\DeclareRobustCommand{\fisheye}{%
  \tikz[baseline=-0.5ex, x=0.25em, y=0.25em, line cap=round, line join=round, scale=0.8]{%
    \useasboundingbox (-0.8,-2.0) rectangle (6.2,2.0);

    \def\FishTop{plot[smooth,tension=0.85] coordinates {(-0.3,0) (0.6,1.1) (2.0,1.6) (3.4,1.1) (4.6,0)}}
    \def\FishBot{plot[smooth,tension=0.85] coordinates {(-0.3,0) (0.6,-1.1) (2.0,-1.6) (3.4,-1.1) (4.6,0)}}

    \begin{scope}
      \path[clip] \FishTop \FishBot -- cycle;
      \draw[line width=1pt] (2.0,0) ellipse [x radius=1.0, y radius=1.85];
    \end{scope}

    \draw[line width=1pt] \FishTop \FishBot;
    \draw[line width=1pt] (4.6,0) -- (5.6,1.2);
    \draw[line width=1pt] (4.6,0) -- (5.6,-1.2);

    \fill (2.0,0) circle (0.42);
  }%
}

\DeclareRobustCommand{\perspcam}[1][1]{%
  \tikz[baseline=0.2ex, x=0.25em, y=0.25em, line cap=round, line join=round, scale=#1]{%
    \useasboundingbox (-0.7,-1.6) rectangle (4.0,1.6);
    \draw[line width=1pt, rounded corners=0.6] (0,0) rectangle (3.4,2.4);
    \draw[line width=1pt, rounded corners=0.4]
      (0.6,2.4) -- (1.4,2.4) -- (1.4,2.9) -- (0.6,2.9) -- cycle;
    \draw[line width=1pt] (2.0,1.2) circle (0.75);
    \draw[line width=1pt] (2.0,1.2) circle (0.20); %
    \fill (2.0,1.2) circle (0.20);
  }%
}

%% file: sec/0_abstract.tex
\begin{abstract}
Radiance fields have emerged as powerful tools for 3D scene reconstruction.
However, casual capture remains challenging due to the narrow field of view of perspective cameras, which limits viewpoint coverage and feature correspondences necessary for reliable camera calibration and reconstruction.
While commercially available 360$^\circ$ cameras offer significantly broader coverage than perspective cameras for the same capture effort, existing 360$^\circ$ reconstruction methods require special capture protocols and pre-processing steps that undermine the promise of radiance fields: {effortless workflows to capture and reconstruct 3D scenes}. 
We propose a practical pipeline for reconstructing 3D scenes directly from raw 360$^\circ$ camera captures.
We require no special capture protocols or pre-processing, and exhibit robustness to a prevalent source of reconstruction errors: the human operator that is visible in all 360$^\circ$ imagery.
To facilitate evaluation, we introduce a multi-tiered dataset of scenes captured as raw dual-fisheye images, establishing a benchmark for robust casual 360$^\circ$ reconstruction.
Our method significantly outperforms not only vanilla 3DGS for 360$^\circ$ cameras but also robust perspective baselines when perspective cameras are simulated from the same capture, demonstrating the advantages of 360$^\circ$ capture for casual reconstruction. Additional results are available at: \url{https://theialab.github.io/fullcircle}
\keywords{360$^\circ$ Reconstruction, Robust Optimization, Casual Capture, Benchmark Dataset}
\end{abstract}

%% file: sec/1_intro.tex
\section{Introduction}
\label{sec:intro}
Recent advances in radiance fields~\cite{nerf, 3dgs} have made photorealistic 3D reconstruction practical, enabling fast training and rendering~\cite{instant_ngp, 3dgs} and deployment in commercial 3D capture products (e.g., Niantic Scaniverse, Polycam, KIRI, and Meta HyperScapes).
This progress has fueled growing efforts to build libraries of  reconstructed scenes~\cite{scannetpp, gaussianworld} and objects~\cite{co3d}.
As these datasets grow and generalizable feed-forward models become a central goal, the bottleneck shifts toward \emph{scalable} capture: collecting 3D-reconstructible data quickly and reliably.

Towards this goal, 360$^\circ$ cameras offer a compelling advantage: each viewpoint observes the
\dr{scene in all directions simultaneously,}
providing not only stronger multi-view constraints for calibration, but also substantially higher visual coverage for reconstruction.
Perspective capture, by contrast, provides only a narrow field of view per frame; consequently, matching the per-image coverage of 360$^\circ$ views demands dense, redundant capture and can require on the order of $10\times$--$30\times$ more images.
Yet most dataset capture pipelines still rely on perspective shots, in part because controlled evidence for 360$^\circ$ capture under matched trajectories remains limited, leaving the gains in reconstruction when using a 360$^\circ$ capture unquantified.

We address this directly with a controlled benchmark comparing a perspective camera capture with a dual-fisheye 360$^\circ$ capture under identical trajectories.
Prior work \cite{camerabench} showed that increasing the field of view improves calibration robustness, and we extend this analysis by demonstrating that 360$^\circ$ cameras further strengthen calibration reliability for 3D reconstruction.
In addition, because 360$^\circ$ images capture substantially more visual information per viewpoint, \textit{when properly processed}, they provide stronger constraints for photometric reconstruction, leading to higher-quality results---particularly for novel views far from the training trajectory; see the left of~\cref{fig:teaser}.

\input{fig/teaser}
A key obstacle remains for reconstructing \emph{human-captured} 360$^\circ$ scenes: the \ul{camera operator} is inevitably visible.
\at{This introduces photometric inconsistencies that not only affect camera estimation, but most importantly reconstruction quality, leading to severe artifacts, as illustrated in the top-right of~\cref{fig:teaser}.}
This is especially true for \emph{casual} handheld capture, which is the low-effort modality needed for scalable data collection.
\at{In a casual capture, the operator may remain static for extended periods of time, causing robust reconstruction methods to fail~\cite{omnilocalrf, robustnerf}.}

\at{One could approach removing the operator from camera footage as an end-to-end segmentation learning problem.
But this would not only require large amounts of training data, but also costly pixel-precise annotations, and the computational resources to train (or at least fine-tune) a video segmentation model at scale.}
\at{We follow an orthogonal approach and instead \textit{exploit} pre-trained vision foundation models, leading to an automatic 360$^\circ$ capture-to-reconstruction pipeline that robustly removes the camera operator.}

We carefully leverage strong pre-trained models in a training-free manner to produce reliable operator masks, and integrate them into a full reconstruction workflow. 
To evaluate robustness, we collect $9$ diverse 360$^\circ$ scenes, each with a dedicated distractor-free test set, and compare against perspective and 360$^\circ$ baselines, including state-of-the-art robust reconstruction methods. 

\at{Beyond delivering a state-of-the-art, ready-to-deploy pipeline for robust 360$^\circ$ reconstruction from casual capture, we aim to enable large-scale 360$^\circ$ data acquisition with automatic and robust operator annotations to support the training of future end-to-end models.
To this end, we openly release both our dataset and capture pipeline, fostering scalable 3D data creation.}

%% file: fig/teaser.tex
\begin{figure}[t]
\begin{center}
\includegraphics[width=.92\linewidth]{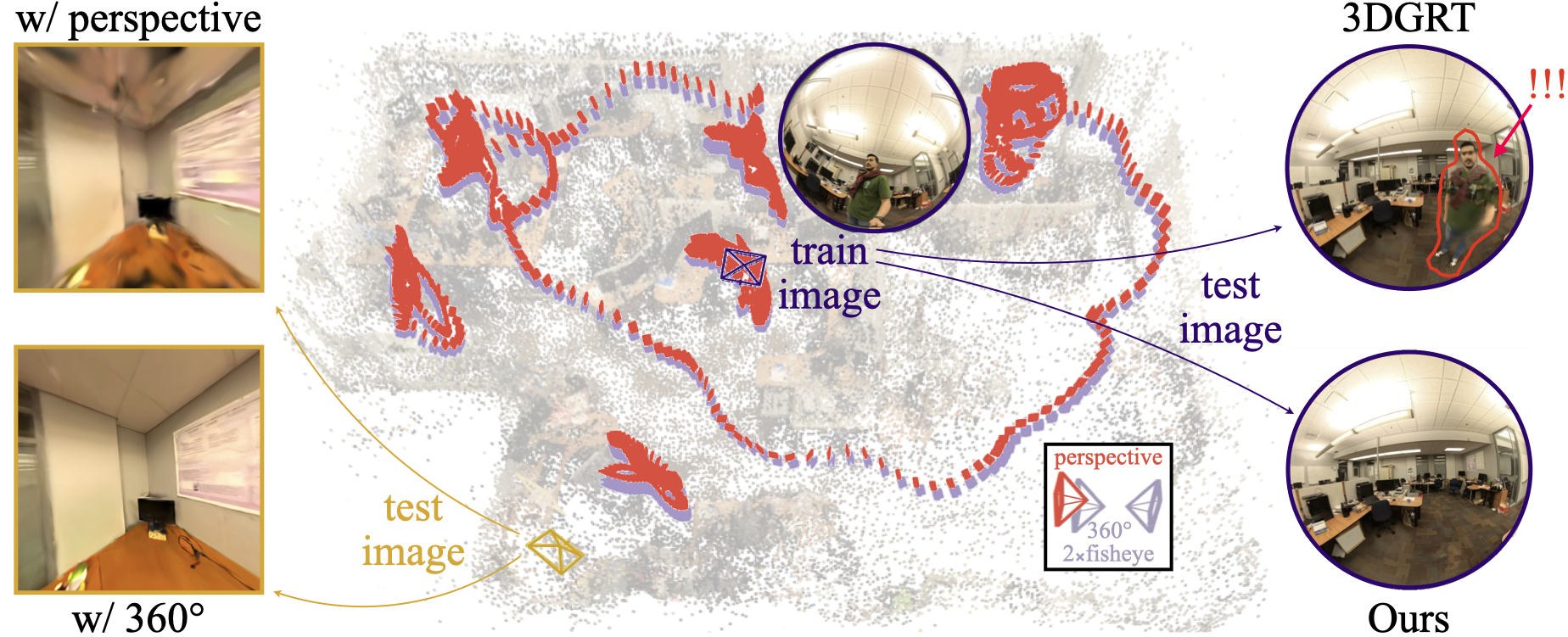}
\end{center}
\captionof{figure}{
\textbf{Teaser} --
With 360$^\circ$ cameras we can recover 3D scenes far more effectively than with traditional perspective cameras, resulting in substantial gains in novel view synthesis (left: perspective vs. 360$^\circ$).
However, casual 360$^\circ$ captures unavoidably include the camera operator, which degrades reconstruction quality if left unaddressed~(e.g.~3DGRT~\cite{3dgrt}).
Our FullCircle method overcomes this challenge, enabling fast, high-quality scene capture using 360$^\circ$ cameras.
}
\label{fig:teaser}
\end{figure}

%% file: sec/2_related.tex
\section{Related Work}
\label{sec:realted}
We review related work on multi-view 3D reconstruction with radiance fields, and then discuss recent works that focus on \textit{non-pinhole} (fisheye, 360$^\circ$) cameras.
We also discuss works to handle \textit{distractors} within a scene.

\paragraph{Radiance fields.}
Multi-view 3D reconstruction has been completely reshaped since the introduction of Neural Radiance Fields (NeRF)~\cite{nerf} and 3D Gaussian Splatting~(3DGS)~\cite{3dgs}, \at{with 3DGS quickly taking over thanks to its much faster rendering, and explicit parameterization.}
Various methods have been proposed to extend radiance fields, including those that enhance efficiency~\cite{instant_ngp, compact3dgs}, robustness to transient distractors~\cite{nerf_in_the_wild, spotlesssplats}, initialization~\cite{3dgsmcmc}, and support for dynamic scenes~\cite{dynamic3dgs, 4dgs}.
In terms of camera compatibility, the original 3DGS is restricted to \textit{perspective} cameras due to its rasterization pipeline.
\at{While NeRF is more flexible thanks to its ray-tracing formulation supporting non-pinhole cameras~\cite{360roam, 360fusionnerf, panogrf, egonerf}, its optimization remains prohibitively slow for practical application.}

\paragraph{Non-pinhole cameras and 3DGS.}
\at{We now focus on 3DGS-based methods capable of optimizing radiance fields from fisheye and omnidirectional images~\cite{360gs, odgs, op43dgs, omnigs}.}
Several works undistort 360$^\circ$ images into cubemap faces for reconstruction and depth estimation~\cite{bifuse, Omnidepth, panoplane360, egocentric}. 
Others adapt the 3DGS rasterization to spherical images by approximating 360$^\circ$ projection with modified perspective rasterization~\cite{360gs, odgs, omnigs, op43dgs, erpgs}.
Conversely, 3D-Gaussian Ray Tracing (3DGRT)~\cite{3dgrt} introduces a ray-tracing formulation for Gaussian primitives, enabling accurate rendering for non-pinhole cameras (e.g.~fisheye and 360$^\circ$).
Gaussian Unscented Transform (3DGUT)~\cite{3dgut} further extends this by introducing unscented transformations, so that it can rasterize 3D Gaussians when lens distortion is present, bringing the ability to not just train, but also render the model without loss of fidelity.
Finally, Seam360GS~\cite{seam360gs} performs seam-aware and exposure-aware 360-degree Gaussian splatting to remove stitching artifacts and photometric inconsistencies in real-world omnidirectional captures.
While the primary focus of these methods is to improve the quality of reconstructions, they still assume \textit{ideal distraction-free} captures.
With 360$^\circ$ cameras, this is never the case, as in casual captures the person holding the camera is \textit{unavoidably} in view.

\paragraph{Reconstructing with distractors.}
In non-controlled environments, reconstruction with NeRFs and 3DGS \textit{must} account for distractors.
In the wild, capturing a perfectly static scene is challenging, often resulting in captures with transient distractors such as moving pedestrians and cars that produce artifacts in the output.
\at{Robustness to transient distractors has been explored for NeRF~\cite{nerf_in_the_wild, robustnerf, nerf-hugs}, and later extended to 3DGS~\cite{swag, gs-w, we-gs, wild3dgs, spotlesssplats}.}
\lily{In our work, we focus on the 360$^\circ$ capture setting, where the most common and impactful distractor is the \ul{camera operator}.}
\at{In this domain the most relevant work is OmniLocalRF~\cite{omnilocalrf}, which proposes an unsupervised approach that separates distractors by leveraging high rendering errors and 3D inconsistencies over time.
However, we find these cues to only work reliably in controlled sequences where the capturer must always remain on the move.
They can break down in casual handheld videos where the operator is near-static for extended periods, such as when remaining stationary to densely image an object of interest within the scene.
By targeting this practical 360$^\circ$ capture regime, we tailor our method toward a more robust capture-to-reconstruction pipeline.}

\paragraph{Segmentation for distorted images.}
\at{Segmentation in omnidirectional images has been an active area of research with approaches relying on domain adaptation techniques to transfer knowledge from perspective images to the omnidirectional domain~\cite{densepass, DPPASS, Trans4PASS, 360fuda, 360fuda++}. Building on the success of Segment Anything Model 2 (SAMv2)\cite{sam} on pinhole images, several works extend SAM-style segmentation to omnidirectional imagery via domain adaptation, typically extracting overlapping patches and treating them as a video input to SAM, with additional training\cite{goodsam} or fine-tuning~\cite{omnisam} for the adaptation.
For segmenting the camera operator in handheld 360$^\circ$ capture, we instead avoid costly adaptation and propose a lightweight strategy that applies SAMv2 ``off-the-shelf''.
Specifically, we use geometric transformations to restrict inference to the low-distortion central region of fisheye frames.
In our experiments, this approach proves more robust than adapted SAMv2 baselines for operator segmentation.
It reduces failures in heavily distorted regions, where segmentation errors lead to reconstruction artifacts, while eliminating the need for task-specific data collection and retraining.}

\paragraph{Dataset collection efforts.}
Radiance-field reconstruction datasets range from controlled captures like ScanNet++~\cite{scannetpp}, designed for accurate calibration and reconstruction, to casual datasets using handheld perspective videos~\cite{robustnerf, nerf_on_the_go, romo, phototourism}.
Existing 360$^\circ$ datasets such as 360Loc~\cite{360loc}, 360VOT~\cite{360vot}, ODIN~\cite{odin}, EgoNeRF~\cite{egonerf}, 360Roam~\cite{360roam}, FIORD~\cite{fiord}, and Ricoh360~\cite{egonerf} employ controlled capture setups or provide pre-stitched panoramas.
To our knowledge, no public dataset offers casual, handheld 360$^\circ$ captures from dual-fisheye cameras; we therefore collect one across diverse scenes to facilitate analysis and benchmarking of reconstruction from casual 360$^\circ$ captures.

%% file: sec/3_method.tex
\section{Method}
\label{sec:method}
Narrow field of view capture imposes multiple constraints, which the user needs to respect to acquire high-quality input data for training radiance fields~\cite{how_to_capture_radiance_fields}.
For example, according to the COLMAP tutorial~\cite{colmap_tutorial}:
\begin{enumerate*}[label=(\roman*)]
    \item texture-less images (e.g. a white wall) should be avoided, as well as 
    \item specularities introduced by shiny surfaces, 
    \item images should be taken to ensure high visual overlap, with each object observed in at least three views, and
    \item enough images should be taken from a relatively similar viewpoint, and yet near duplicates and pure camera rotations should be avoided~(i.e., take a few steps after each photograph is taken).
\end{enumerate*}
Clearly, such constraints are impractical for casual users. 
In contrast, professional reconstruction tools (e.g.~Meta Horizon Hyperscape ~\cite{MetaHyperscapeCaptureBeta_2025, MetaQuestHelp_instructions_2025}), often incorporate AR-based guidance systems that actively assist users in maintaining coverage, overlap, and scene diversity during capture. 

\paragraph{Data capture.}
In contrast to a traditional radiance field capture and reconstruction process, which depends on thousands of manually placed perspective photographs, our capture pipeline is built on a \ul{casual video} capture from a 360$^\circ$ camera.
In order for the process to be employed at scale by \ul{non-expert users}, our pipeline must not require any tedious, methodical camera placement, and must be robust to a variety of behaviors from the person performing the capture, including cases where the capturer does not actively cover all viewing directions, as well as extended periods where they may remain stationary, such as while capturing close-up images of a small area of interest in high details.
These requirements present unique challenges in the context of 360$^\circ$ capture, as the \textit{capturer} will be visible at all times, and must be removed from the final scene reconstruction.
\input{tab/stats}

\paragraph{Dataset.}
To evaluate our technique for casual capture, we capture a dataset of 9 scenes with a single consumer-grade Insta360 X4 360$^\circ$~(dual-fisheye) camera, readily available through mainstream retail channels, to reflect its widespread adoption among casual users.
During these captures, we \ul{emulate non-expert capturers' behaviors}, including both periods of motion where the person actively moves around the scene, as well as periods where the person remains static, and only moves the camera.
For statistics on the lengths of these captures, including the static and dynamic parts, as well as other details, please refer to~\cref{tab:stats}.
For the purpose of quantitative evaluation, we also capture a \ul{golden test set} of images for each scene using a \textit{tripod}, where no people are visible, and the visible parts of the tripod are masked out.
\footnote{{This capture procedure mirrors an expert capture session, which requires much longer capture time, and deliberate choices during capture.}} 
This enables unbiased measurement of the novel view quality from views which are not sampled from the training trajectory, and which include areas that were occluded by the capturer during the video.

\begin{figure*}[t!]
\centering
\input{fig/stitch}
\input{fig/masks}
\end{figure*}

\paragraph{Raw fisheye input.}
The manufacturer-provided software with the 360$^\circ$ camera by default outputs \textit{stitched} omnidirectional/panorama images in \textit{equirectangular} projection; see~\cref{fig:stitch} (left).
However, using these stitched images directly for training is problematic due to the inevitable \textit{optical misalignment} between the two original fisheye views in consumer cameras.
This misalignment causes artifacts around the stitching boundary, and thus results in lower quality reconstructions when used as ground truth; see~\cref{fig:stitch}. 
Instead, we extract the original (un-stitched) fisheye frames, and use these as training inputs.
While these frames do not show any artifacts from stitching, they do show \textit{distortion} and \textit{chromatic aberration} effects around the outer boundary of the camera's field of view.
We mask out these pixels with a boundary mask,
before continuing to camera estimation and reconstruction; see~\cref{fig:masks}.

\subsection{Capturer mask estimation}
\label{sec:maskpipeline}
Because the training views include the person holding the camera during capture, we generate masks to remove the person before proceeding to reconstruction. Rather than treating the presence of the capturer as a nuisance, we take advantage of this consistency and propose a simple, robust masking strategy that leverages the fact that the capturer \textit{appears in all views}, and \textit{moves continuously} through time.
This stands in contrast to previous perspective-robust methods that rely on analyzing photometric error during reconstruction to infer masks~\cite{spotlesssplats, wildgaussians, robustnerf, omnilocalrf}.
{To identify pixels belonging to the capturer, we employ a two-stage masking process.
First, we estimate the approximate location of the person, and then we generate a synthetic fisheye view centered on them which is used to robustly estimate the final mask.}
The entire process is illustrated in \cref{fig:maskpipeline}.

\input{fig/maskpipeline}

\paragraph{Finding the capturer.}
We start by extracting a set of 16 overlapping virtual 90$^\circ$ pinhole camera views that cover the sphere.
Independently for each of these frames, a mask is predicted using SAMv2~\cite{sam} at the location of the capturer in the frame predicted by YOLOv8~\cite{yolo} with a ``person'' prompt.\footnote{This procedure relies on the person detector correctly localizing the capturer in at least a few frames, with minimal false positives, and on the capturer being the \textit{dominant} human distractor present in most frames.}
An approximate global direction for the capturer is then obtained as the \textit{average direction} of every pixel in each pinhole frame that was classified as part of the capturer.
Synthetic 180$^\circ$ fisheye frames centered on this direction are then generated from the omnidirectional images.

\paragraph{Segmenting capturer (through time).}
After synthesizing fisheye images such that the capturer remains centered, we automatically prompt SAMv2~\cite{sam} by selecting the image center as the prompt location.
This setup allows us to exploit temporal consistency, as mask propagation through time becomes robust when the capturer stays near the image center.
We further leverage the observation that auto-stitching artifacts, though detrimental to 3DGS reconstruction, remain largely tolerable for pre-trained video segmentation models like SAMv2~\cite{sam}.
To account for uncertainties in the segmenter predictions, we also dilate the predicted final masks by a negligible amount ({${\approx}$4} pixels).
After intermediate conversions between omnidirectional, synthetic fisheye, and cubemap representations, the final capturer masks are propagated back to the original fisheye domain, with all RGB data preserved in their raw form throughout the process.

\subsection{Training the radiance field}
After having masked input fisheye images, we continue with a mostly standard 3DGS training process.
We estimate camera poses with COLMAP~\cite{colmap1, colmap2}, using raw front and back fisheye images, and generate a SfM point cloud to initialize 3DGS training.
The composed masks are used in both steps to avoid failures due to the presence of the capturer in the training images; see~\cref{sec:fisheye_capture} and~\cref{tab:stats}.
Classical 3DGS training undistorts fisheye images with COLMAP before training.
Conversely, we build upon 3D Gaussian Ray Tracing~(3DGRT)~\cite{3dgrt}, which is capable of rendering non-pinhole images, and instead use the masked fisheye frames for training.
Importantly, we show that using raw fisheye images with 3DGRT achieves higher reconstruction quality compared to training a 3DGS model on undistorted images, due to its higher coverage; see~\cref{fig:comparison-p}.
3DGRT model is trained on all views from the camera trajectory which were successfully estimated by COLMAP, with separately captured tripod views used as test set.

%% file: tab/stats.tex
\begin{table}[!t]
\caption{
We collect a dataset of dual-fisheye scene captures with different levels of difficulty. 
We report the number of frames where the capturer is static/dynamic, and the number of COLMAP pose failures (w/ and w/o transient masks). The full image set is released to support future work on robust calibration.
}
\vspace*{-1em}
\label{tab:stats}
\centering
\setlength{\tabcolsep}{16pt}
\resizebox{\columnwidth}{!}{
\begin{tabular}{@{}lccccccr@{}}
\textbf{Scene} &
\textbf{\#training} &
\textbf{\#test} &
\textbf{\#total} &
\textbf{\#w/ mask} &
\textbf{{\#w/o mask}} &
\textbf{duration}&
\textbf{difficulty} \\
\midrule
Room 1  & 442 / 117 & 38 & 597 & --  &   8 & 88s / 23s & Easy   \\
Flat 1  & 440 / 153 & 46 & 639 & --  &  56 & 88s / 31s & Easy   \\
Flat 2  & 434 / 107 & 26 & 567 & --  &   1 & 87s / 21s & Easy   \\
Room 2  & 252 /~~82 & 28 & 362 & --  &  -- & 50s / 16s & Easy   \\
Room 3  & 257 / 100 & 30 & 387 &  1  & 104 & 51s / 20s & Medium \\
Lab     & 431 / 116 & 34 & 581 & --  &   3 & 86s / 23s & Medium \\
Lounge  & 413 / 120 & 22 & 555 & --  &  -- & 83s / 24s & Medium \\
Persons & 448 / 129 & 34 & 611 & 110 & 611 & 90s / 26s & Hard   \\
Dark    & 434 / 113 & 35 & 582 & --  &   8 & 87s / 23s & Hard   \\

\end{tabular}
}
\vspace*{-1em}
\end{table}

%% file: fig/stitch.tex
\begin{minipage}[t]{0.49\linewidth}

\centering

\includegraphics[width=\linewidth]{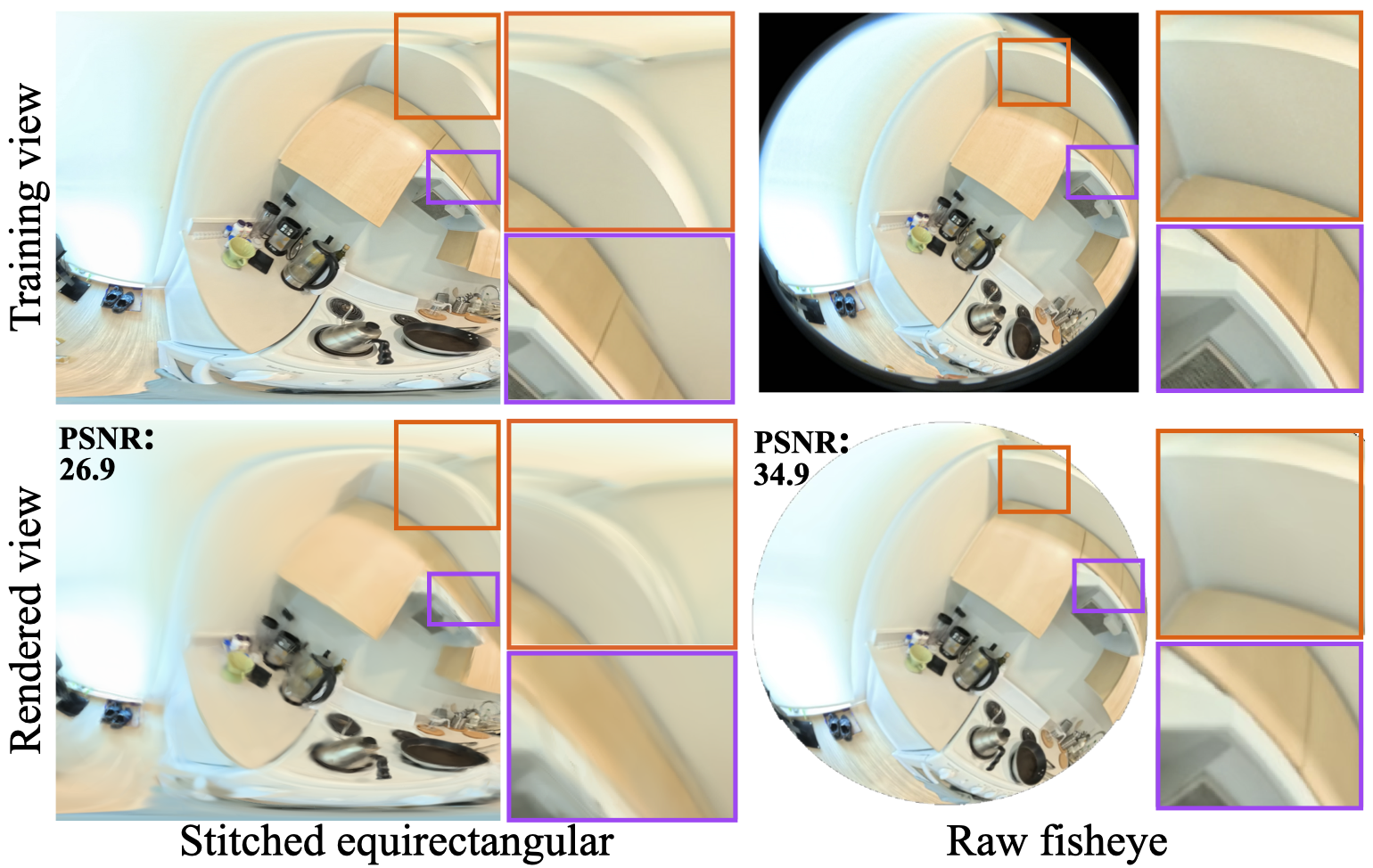}
\captionof{figure}{\textbf{Stitching artifacts} -- 
Equirectangular inputs contain stitching artifacts (top left) that lead to noisy edges and reduced PSNR in the reconstruction (bottom left). Our method, trained on raw fisheye images, avoids these artifacts (right).
}
\label{fig:stitch}
\end{minipage}

%% file: fig/masks.tex
\begin{minipage}[t]{0.49\linewidth}
\centering
\includegraphics[width=\linewidth]{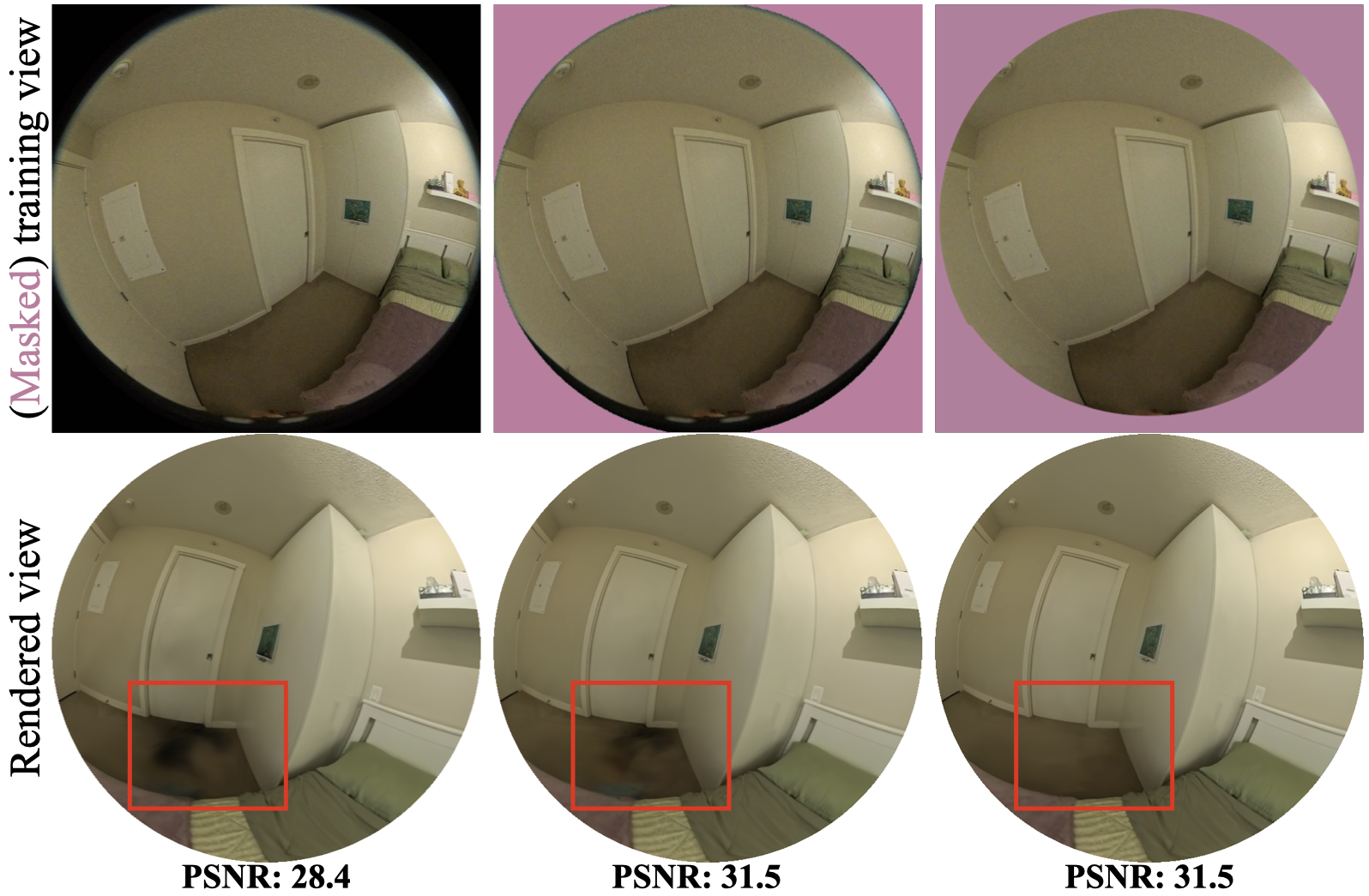}
\captionof{figure}{\textbf{Boundary masking} -- 
Unmasked black boundary pixels contaminate the reconstruction (left). Minimal masking (\textcolor{mypink}{pink}) leaves blue artifacts from edge distortion and color aberration (middle). Our dilated mask prevents artifacts (right).}
\label{fig:masks}
\end{minipage}

%% file: fig/maskpipeline.tex
\begin{figure*}[t!]
\centering
\includegraphics[width=\linewidth]{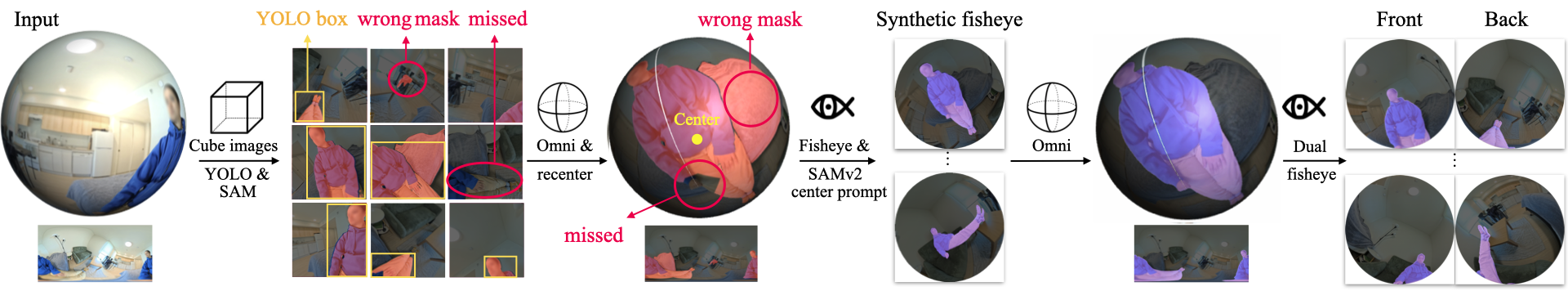}

\captionof{figure}{\textbf{Capturer mask estimation} -- 
Our masking pipeline first detects the capturer using YOLOv8~\cite{yolo} and SAMv2~\cite{sam} (\textcolor{lightred}{red} masks). 
These initial masks roughly localize the capturer but include missing or incorrect regions that can degrade reconstruction. 
We then re-center the omni images on the detected capturer, render synthetic fisheyes, and re-run SAMv2 with a center-point prompt to obtain refined, temporally consistent masks (\textcolor{darkpurple}{purple} masks), which are mapped back to the original dual-fisheye inputs.
}
\label{fig:maskpipeline}
\end{figure*}

%% file: sec/4_results.tex
\section{Experiments}
We first show the effectiveness of 360$^\circ$ capture compared to conventional perspective capture under similar camera trajectories~(\cref{sec:fisheye_capture}).
We show that 360$^\circ$ cameras are inherently more suitable for casual scene acquisition due to their wider coverage and consequently easier camera calibration.
Next, we discuss why undistorting 360$^\circ$ images before reconstruction is suboptimal, showing using raw fisheye inputs yields higher-fidelity reconstructions~(\cref{sec:pers_comp}).
We then show that, when operating directly on raw fisheye images, our method effectively suppresses the distractor capturer and reconstructs high-quality scenes, outperforming baselines designed for fisheye inputs~(\cref{sec:fisheye_comp}).
Finally, we ablate our pipeline design and compare our capturer segmentation method to a SOTA omnidirectional image segmentation method~(\cref{sec:ablation}).

\paragraph{Implementation details.}
Our implementation is based on 3DGRT~\cite{3dgrt}, trained with the Adam optimizer and default learning rates for 30k iterations.
3DGS perspective baselines follow their official implementations, and NeRF baselines use the Nerfacto~\cite{nerfacto} model, with robust extensions and different camera models integrated. 
Camera calibration (for both fisheye and undistorted fisheye) and image undistortion are performed using COLMAP v3.12~\cite{colmap1, colmap2}. \yf{For omnidirectional images, camera poses are estimated using SphereSfM~\cite{spheresfm}.} Unless otherwise specified, we use images down-sampled by a factor of $4\times$ for faster processing and reduced memory usage.
Human detection is done with YOLOv8s~\cite{yolo}, and segmentation with mask propagation in SAMv2.1~\cite{sam}~(large hierarchical backbone). %

\subsection{Dual-fisheye vs. perspective --~\cref{fig:why360}}
\label{sec:fisheye_capture}
One of our main claims is that 360$^\circ$ dual-fisheye cameras provide a more suitable setup for casual data capture than conventional perspective cameras for 3D reconstruction. 
To illustrate this, we conduct a controlled experiment comparing camera calibration as well as reconstruction quality between fisheye and perspective frames.
The images are captured along \textit{identical} camera trajectories to emulate two comparable 80-second casual video captures by a non-expert user.

\paragraph{Data.}
We collect a dataset of 191 dual-fisheye images corresponding to approximately {80} seconds of casual, handheld capture using a dual-fisheye camera.
However, the camera is \textit{carefully} operated like a classical perspective camera, always directed toward scene structures of interest (there are no human distractors in the front fisheye view).
To ensure perfectly consistent exposure and color balance between fisheye and perspective settings, we do not perform a separate perspective capture; instead, in this controlled experiment, we use undistorted images from the front fisheye to simulate perspective views.

\input{fig/why360}

\paragraph{Evaluation variants.}
We evaluate three different settings:
\begin{itemize}
    \item (\fisheye–\fisheye): both calibration and reconstruction are performed directly on the raw dual-fisheye images using COLMAP followed by our robust reconstruction pipeline, with no undistortion preprocessing.
    \item (\fisheye–\perspcam): camera calibration is performed on the raw dual-fisheye images, but reconstruction uses only the undistorted perspective images from the front camera.
    \item (\perspcam–\perspcam): calibration and reconstruction are performed on the front camera's undistorted perspective images.
\end{itemize}

\paragraph{Metrics.}
We assess both calibration and reconstruction quality through novel view synthesis performance, reporting PSNR on full-resolution images of $2880\times2880$ as a quantitative image quality metric.
We assess the reconstruction quality at novel views, outside the distribution of training camera views. We report PSNR, SSIM and LPIPS.

\paragraph{Analysis.}
As shown in \cref{fig:why360}, all three pipelines produce high-quality reconstructions near the training trajectory, but performance diverges as the viewpoint moves toward out-of-distribution test views. 
The (\fisheye–\fisheye) pipeline maintains stable quality across this trajectory, while (\fisheye–\perspcam) and (\perspcam–\perspcam) degrade significantly.
Notably, (\fisheye–\perspcam) consistently outperforms (\perspcam–\perspcam), even though the same set of undistorted perspective images is used for reconstruction. 
This indicates that calibration performed on raw dual-fisheye images is more accurate and has a direct impact on 3D reconstruction robustness.
{Overall, this experiment confirms that, if we were able to remove the capturer from input images, fisheye images are \textit{significantly} more effective than perspective images in reconstructing scenes with high fidelity (better calibration, and better scene coverage).
We now, therefore, shift our focus to dual-fisheye data.

\input{fig/comparison-p}
\subsection{Reconstruction via undistortion --~\cref{fig:comparison-p}}
\label{sec:pers_comp}
While performing capture and calibration in the dual-fisheye domain offers the benefits shown in the previous section, reconstruction from such images can be approached in different ways. A common strategy is to first undistort the front and back fisheye views and then perform reconstruction on the resulting perspective images. This conveniently enables the use of robust perspective-based methods~\cite{spotlesssplats, nerf_on_the_go} designed to handle distractors. To highlight the advantages of reconstructing directly from dual-fisheye inputs, we compare our pipeline, which is trained on raw fisheye images from our dataset, against perspective baselines trained on undistorted front and back views. We further evaluate the robustness of these perspective baselines to the transient capturer, compared to our method.

\paragraph{Dataset.}
Due to the lack of publicly available dual-fisheye datasets captured casually with the human capturer visible, we collected our own dataset of nine scenes (eight indoor and one outdoor) spanning a range of reconstruction difficulties. Detailed statistics are provided in~\cref{tab:stats}. 
The videos are recorded at 5 FPS and a resolution of $2880 \times 2880$ pixels using an Insta360 X4 dual-fisheye camera.
Easy scenes correspond to typical indoor captures, medium scenes include view-dependent effects, specular surfaces, and long-horizon outdoor environments, while hard scenes feature \yf{challenging illumination} and multiple human distractors.
The dataset will be publicly released to support future research.
For experiments in this section, the front and back fisheye images are undistorted using COLMAP~\cite{colmap1,colmap2}, and then used as input data to reconstruction. All baselines use poses estimated from the dual-fisheye images using COLMAP. 

\paragraph{Metrics.}
We report PSNR as the main novel-view synthesis metric on perspective renderings from the test camera views. 
Test fisheye images are captured from a tripod to minimize distractor interference. The tripod body is masked out when computing evaluation metrics.
SSIM and LPIPS are included in the supplementary material.

\input{fig/comparison-f}
\paragraph{Baselines.}
We compare against 3DGS~\cite{3dgs} and its robust variant SpotLessSplats~\cite{spotlesssplats} for perspective-based Gaussian Splatting baselines. From SLS, we use its MLP variant trained with Stable Diffusion features for distractor segmentation. We also provide comparison against NeRF On-the-go (NOTG)~\cite{nerf_on_the_go} as the state-of-the-art robust NeRF method. 
We re-implement NOTG on top of Nerfacto~\cite{nerfacto} for faster training and rendering.
All methods are trained on undistorted front and back images, except ours which is trained on raw fisheye images.

\paragraph{Analysis.}
3DGS is a non-robust baseline and frequently reconstructs the human capturer as noisy distractor artifacts.
SLS-MLP~\cite{spotlesssplats} improves robustness to distractors but often exhibits visible artifacts due to reduced coverage and distortions introduced during the undistortion and resampling phase (\cref{fig:comparison-p}, second row).
In scenes where the capturer remains static for extended periods, SLS also tends to partially reconstruct the distractor itself (\cref{fig:comparison-p}, first row).
In contrast, our method produces clean, high-fidelity reconstructions without visible distractor artifacts and outperforms baselines quantitatively across most scenes, with the exception of one scene with an extreme illumination setting.

\subsection{Dual-fisheye reconstruction --~\cref{fig:comparison-f}}

\label{sec:fisheye_comp}
Following~\cref{sec:pers_comp}, we focus on methods capable of reconstructing directly from fisheye images without undistortion preprocessing and achieving higher reconstruction fidelity.
We evaluate our 3D reconstruction pipeline that works on raw dual-fisheye casual captures, on our collected dataset, where the capturer is present in all frames, either in the front camera, the back camera, or both, emulating a non-expert capture. We compare our method to other robust and non-robust methods that operate on 360$^\circ$ imagery.

\paragraph{Dataset.}
We use the same dataset introduced in~\cref{sec:pers_comp}, following the same train/test split. All images are calibrated using COLMAP on dual-fisheye inputs.

\paragraph{Metrics.}
We evaluate all methods on clean test fisheye images captured from a tripod, with the tripod body masked out. 
PSNR results are reported in~\cref{fig:comparison-f}. SSIM and LPIPS are included in the supplementary material.

\paragraph{Baselines.}
We compare against 3DGRT~\cite{3dgrt}: a non-robust Gaussian Splatting baseline that supports fisheye cameras. 
For a robust version, we augment 3DGRT with SpotLessSplats~\cite{spotlesssplats} SLS-MLP variant.
Additionally, we include NeRF-based robust baselines.
\yf{Specifically, we use OmniLocalRF (OLRF)~\cite{omnilocalrf} on omnidirectional images stitched from the raw fisheye frames, and calculate metrics on the rendered omnidirectional images}. When multiple videos are available in one scene, the frames are appended to create a continuous video as input to OmniLocalRF. Further, for a fair comparison, we provide SphereSfM camera poses to OmniLocalRF instead of optimizing them.
We re-implemented NeRF On-the-go (NOTG)~\cite{nerf_on_the_go} on Nerfacto~\cite{nerfacto} base model for faster convergence and rendering.
In all methods, the outer ring of the fisheye images is masked out.

\paragraph{Analysis.}
Our method \yf{\ul{outperforms all fisheye-compatible and also omnidirectional baselines}} both qualitatively and quantitatively.
The vanilla 3DGRT baseline~\cite{3dgrt} fails to handle dynamic distractors, frequently reconstructing the human capturer as noisy artifacts. 
Adding the SLS robust loss~\cite{spotlesssplats} improves results, but struggles when the capturer remains momentarily immobile, failing to separate them from the static background and leaking into reconstruction early on in training.
\yf{OLRF often reconstructs the camera holder when they remain mostly static, failing to distinguish them from the static scene.}
NOTG often overestimates uncertainty maps when paired with a fisheye camera model, leading to under-reconstruction in regions near the capturer.
In contrast, our method explicitly models the persistent presence of the human capturer across frames and segments them out, producing clean, high-fidelity reconstructions even in complex or reflective environments.
For the hard scenes, \yf{our method occasionally reconstructs partial geometry of other—typically static—distractors}; see~\cref{fig:limit}.

\subsection{Ablation study -- \cref{fig:omnisam} \& \cref{fig:synth}}
\label{sec:ablation}

\paragraph{Camera operator segmentation.}
We ablate our operator segmentation design by comparing it against OmniSAM~\cite{omnisam}, a recent general-purpose segmentation method for omnidirectional images. Specifically, we extract the \textit{person}-class predictions from OmniSAM’s 13-class semantic output, map the resulting masks to the raw fisheye frames, and use them to remove distractors in our pipeline.

\input{fig/omnisam}

\input{fig/synth}

We find that applying OmniSAM directly to omnidirectional images is less robust in highly distorted regions, often leaving residual operator pixels that appear as visible artifacts in the reconstruction. Additional qualitative comparisons in the supplementary material show that this failure mode occurs in highly distorted regions when using this general-purpose omnidirectional image segmentor.
In contrast, our approach segments the operator more reliably by re-centering the fisheye views and performing segmentation in a low-distortion region (i.e., when the operator is near the center of the fisheye frames). This allows SAMv2 to observe a minimally deformed operator silhouette. Our method is therefore not only more robust for camera operator segmentation, but also does not require complex and costly retraining or fine-tuning of SAMv2.

\input{fig/limit}

\paragraph{Centering fisheye frame.}
We also ablate the effect of synthesizing fisheye frames centered on the camera operator. As shown in \cref{fig:synth}, without this centering step—i.e., when using YOLOv8 and SAMv2 output masks on overlapping patches directly—the operator is not segmented consistently, leading to partial reconstruction of the camera holder in the renderings.

%% file: fig/why360.tex
\begin{figure*}[t]
\centering
\includegraphics[width=\linewidth]{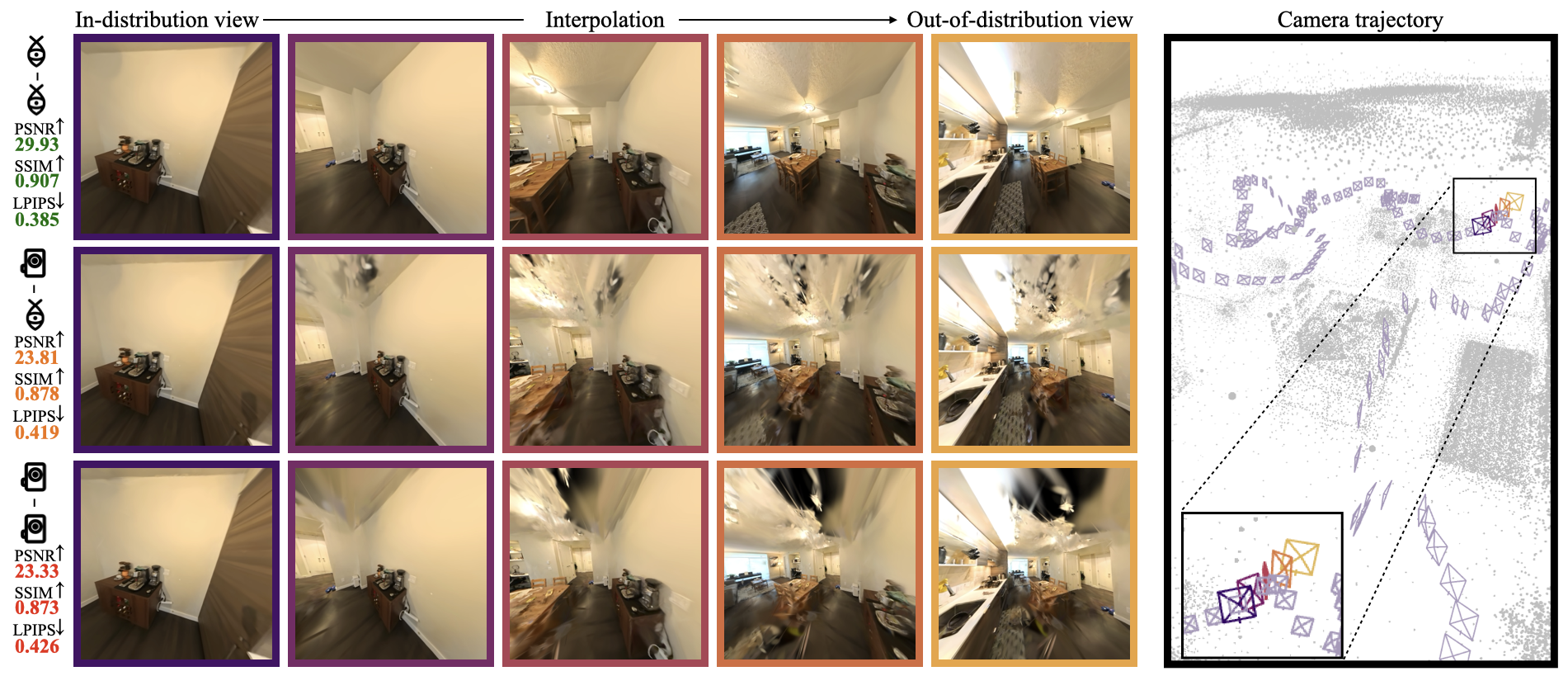}
\captionof{figure}{
\textbf{Dual-fisheye vs. perspective capture} --  Using dual-fisheye images for both calibration and reconstruction (\fisheye–\fisheye) yields higher-quality novel views and stable geometry than when reconstruction~(\fisheye–\perspcam), or both calibration and reconstruction~(\perspcam–\perspcam), relies on perspective images.
This advantage becomes most apparent when moving beyond the in-distribution training trajectory (\textcolor{darkpurple}{purple} cameras) toward out-of-distribution test views (\textcolor{darkyellow}{yellow} camera), where (\fisheye–\perspcam) and (\perspcam–\perspcam) exhibit a gradual degradation in reconstruction quality.
}
\label{fig:why360}
\vspace{-0.4cm}
\end{figure*}

%% file: fig/comparison-p.tex
\begin{figure*}[t]
\centering
\includegraphics[width=1.0\textwidth]{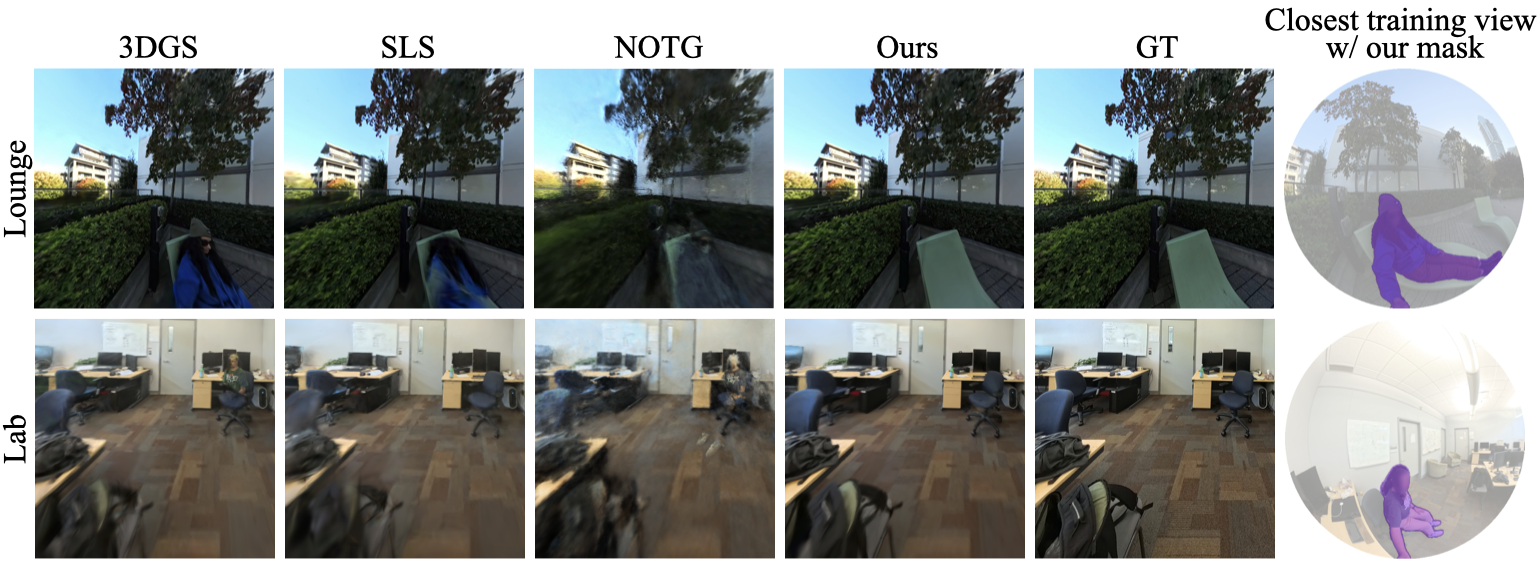}
\setlength{\tabcolsep}{8pt}
\resizebox{\textwidth}{!}{
\begin{tabular}{lc|ccccccccc|c}
\textbf{Method} & Robust & Room 1 & Flat 1 & Flat 2 & Room 2 & Room 3 & Lab & Lounge & Persons & Dark & \textbf{Mean}\\
\midrule
3DGS~\cite{3dgs}              & \xmark & 26.8 & 24.3 & 23.3 & 25.7 & 23.9 & 24.6 & 20.6 & 24.7 & 24.1          & 24.2\\
SLS-MLP~\cite{spotlesssplats} & \cmark & 29.9 & 25.2 & 26.3 & 28.4 & 25.5 & 25.6 & 20.8 & 26.9 & \textbf{28.2} & 26.3\\
NOTG~\cite{nerf_on_the_go}    & \cmark & 25.1 & 20.3 & 23.4 & 22.4 & 21.9 & 22.5 & 20.1 & 26.3 & 24.3          & 22.9\\

\midrule
\textbf{FullCircle (Ours)}    & \cmark & \textbf{31.6} & \textbf{26.9} & \textbf{28.4} & \textbf{30.6} & \textbf{28.3} & \textbf{27.8} & \textbf{22.5} & \textbf{28.9} & 27.2 & \textbf{28.0}\\
\end{tabular}
}
\caption{
\textbf{Effect of undistortion} -- Using fisheye frames with \methodname yields higher-quality reconstructions, both qualitatively and quantitatively, compared to perspective baselines—robust and non-robust—that operate on preprocessed, undistorted images.
}
\vspace{-.5em}
\label{fig:comparison-p}
\end{figure*}

%% file: fig/comparison-f.tex
\begin{figure*}[t]
\centering
\includegraphics[width=1.0\textwidth]{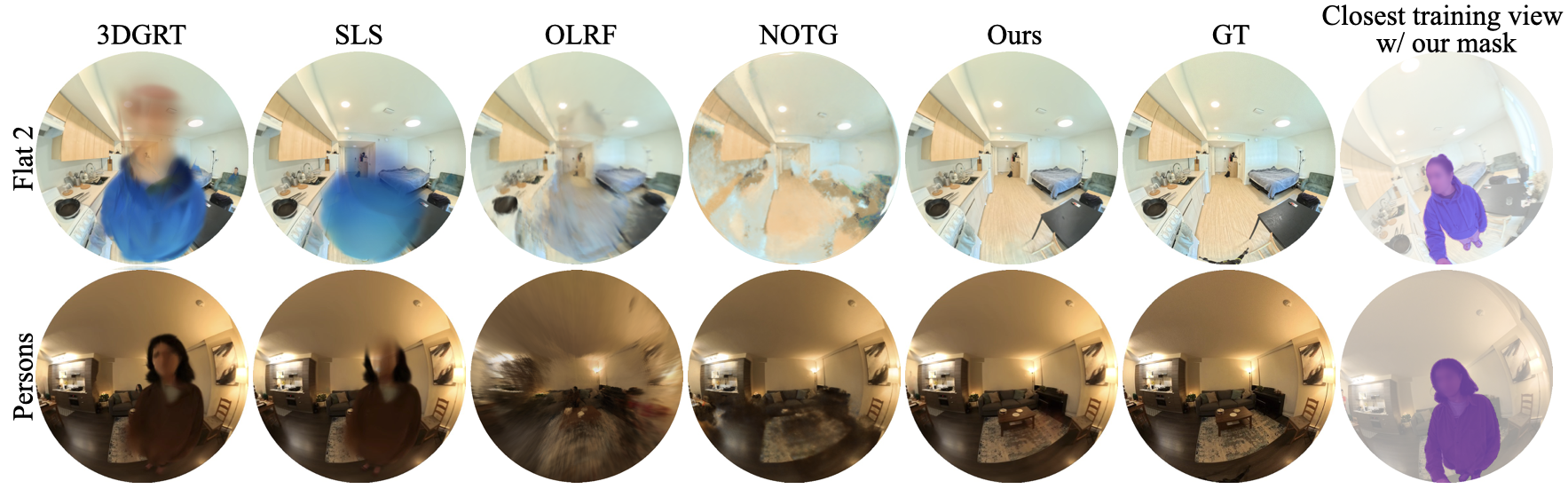}
\setlength{\tabcolsep}{8pt}
\resizebox{\textwidth}{!}{
\begin{tabular}{lc|ccccccccc|c}
\textbf{Method} & Robust & Room 1 & Flat 1 & Flat 2 & Room 2 & Room 3 & Lab & Lounge & Persons & Dark & \textbf{Mean} \\
\midrule
3DGRT~\cite{3dgrt}             & \xmark & 27.4 & 26.8 & 25.1 & 27.9 & 28.4 & 26.7 & 23.4 & 27.6 & 25.7 & 26.5\\
SLS-MLP~\cite{spotlesssplats}  & \cmark & 29.5 & 28.7 & 28.0 & 31.3 & 29.1 & 28.6 & 23.7 & 29.1 & 26.7 & 28.3\\
OLRF~\cite{omnilocalrf} & \cmark & 21.0 & 19.6 & 20.9 & 23.9 & 20.7 & 17.9 & 18.4 & 20.9 & 19.9 & 20.3\\
NOTG~\cite{nerf_on_the_go}     & \cmark & 24.2 & 19.4 & 19.2 & 27.7 & 22.1 & 17.4 & 19.8 & 24.6 & 20.9 & 21.7\\
\midrule
\textbf{FullCircle (Ours)} & \cmark & \textbf{32.6} & \textbf{29.1} & \textbf{30.5 }& \textbf{31.6} & \textbf{29.9} & \textbf{29.5} & \textbf{24.4} & \textbf{31.0} & \textbf{ 27.8} & \textbf{29.6}\\
\end{tabular}
}
\caption{
\textbf{Fisheye reconstruction} -- Our method reliably removes human distractors and reconstructs high-quality 3D scenes. The table reports PSNR, where our method outperforms both robust and non-robust radiance-field baselines.
\yf{OLRF renderings are mapped to fisheye frames for qualitative comparison only.}
}
\vspace{-.6cm}
\label{fig:comparison-f}
\end{figure*}

%% file: fig/omnisam.tex
\begin{figure*}[ht!]
\centering
\includegraphics[width=1.0\textwidth]{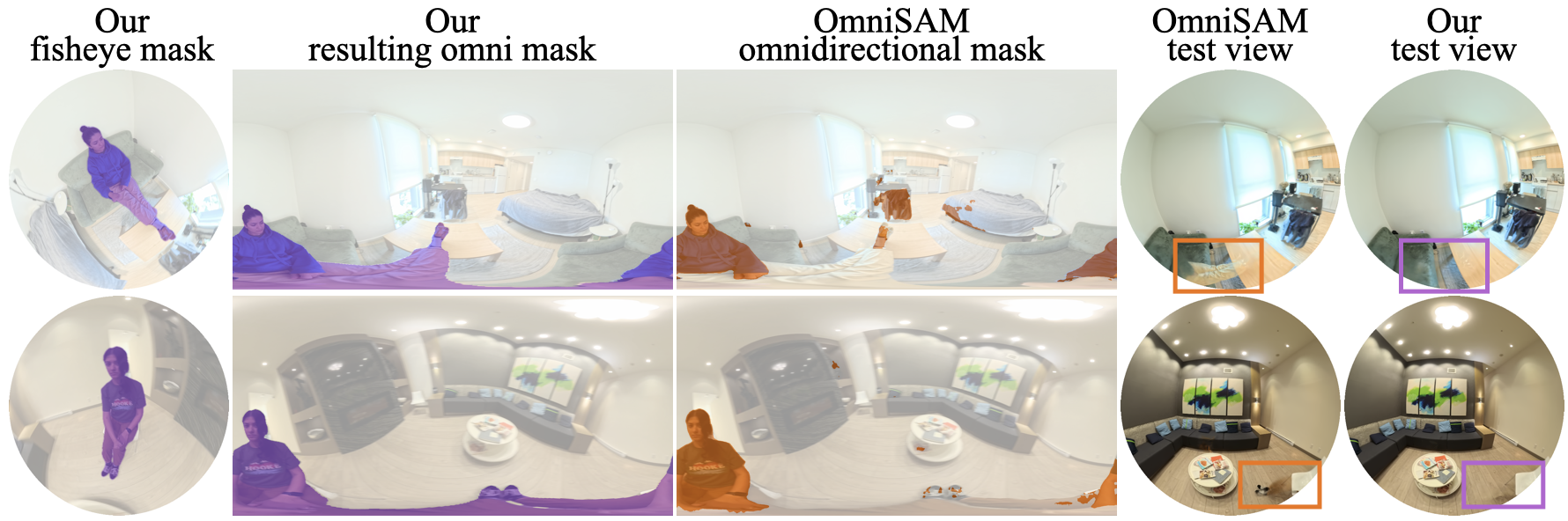}
\setlength{\tabcolsep}{8pt}
\resizebox{\textwidth}{!}{
\begin{tabular}{l|ccccccccc|c}
\textbf{Method} & Room 1 & Flat 1 & Flat 2 & Room 2 & Room 3 & Lab & Lounge & Persons & Dark & \textbf{Mean} \\
\midrule
3DGRT~\cite{3dgrt} + OmniSAM~\cite{omnisam} & 32.1 & 28.5 & 29.1 & 31.1 & 29.5 & 28.5 & 24.3 & 30.8 & 27.4	& 29.0\\
\textbf{FullCircle (Ours)}                  & \textbf{32.6} & \textbf{29.1} & \textbf{30.5} & \textbf{31.6} & \textbf{29.9} & \textbf{29.5} & \textbf{24.4} & \textbf{31.0} & \textbf{27.8} & \textbf{29.6}\\
\end{tabular}
}
\caption{
\textbf{Camera operator segmentation} -- We compare our masking approach for removing the camera operator with OmniSAM, a recent segmentation method for omnidirectional images, both qualitatively and quantitatively.
}
\vspace{-1.2cm}
\label{fig:omnisam}
\end{figure*}

%% file: fig/synth.tex
\begin{figure}[ht!]
\centering
\includegraphics[width=0.85\textwidth]{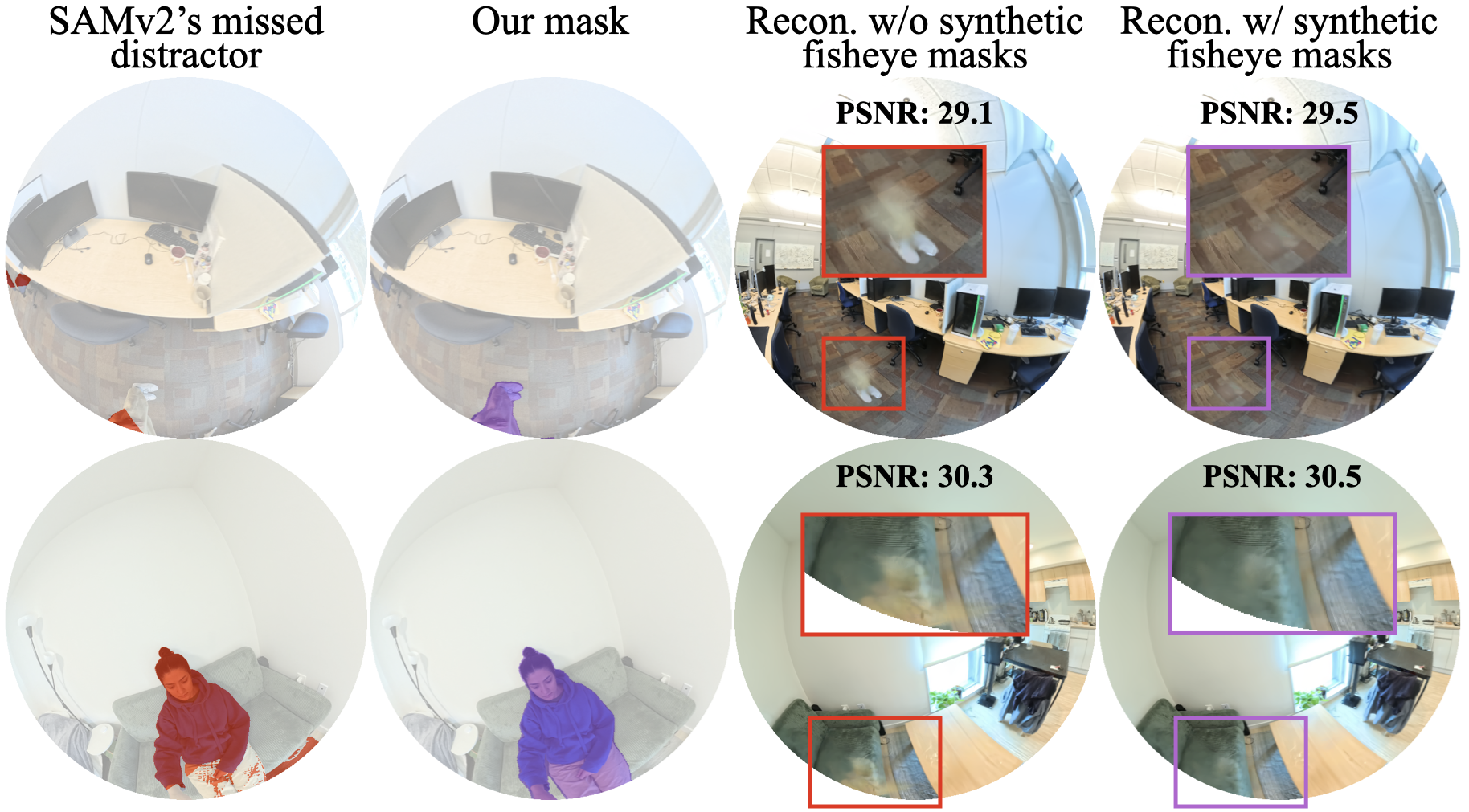}

\captionof{figure}
{\textbf{Synthetic fisheye} -- 
Without re-centering the fisheye images synthetically, SAMv2~\cite{sam} misses the capturer in some frames, resulting in a noisy reconstruction.}
\vspace{-0.3cm}
\label{fig:synth}
\end{figure}

%% file: fig/limit.tex
\begin{figure}[t!]
\centering
\includegraphics[width=0.85\linewidth]{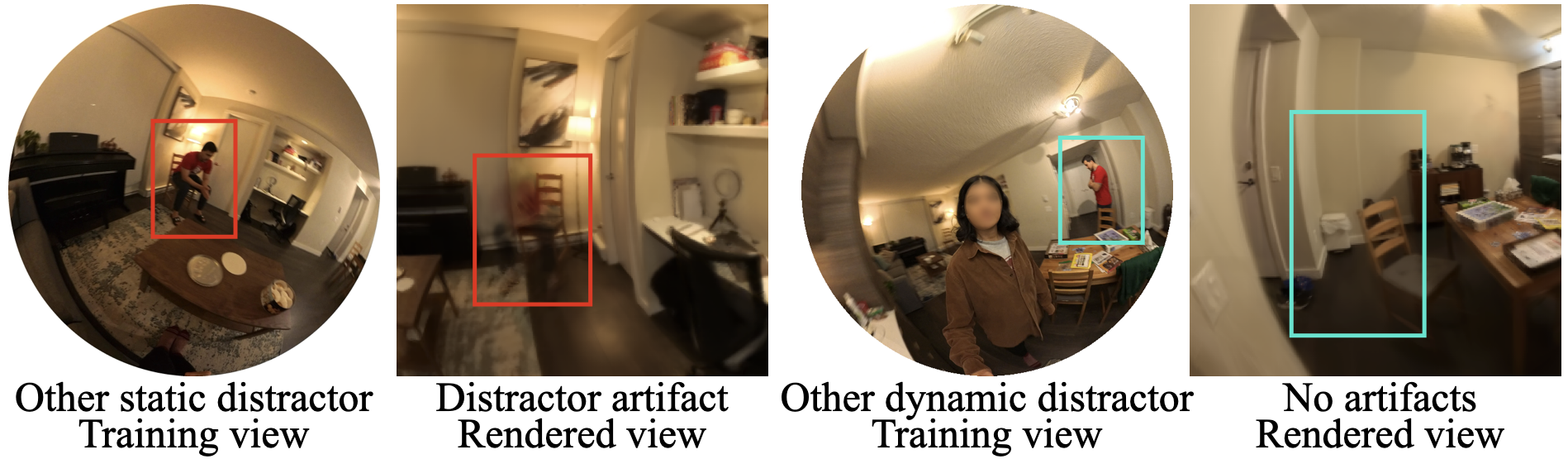}
\captionof{figure}{\textbf{Limitations} -- 
We find that 360$^\circ$ coverage provides enough data redundancy to allow our pipeline to robustly handle other moving distractors (right), but our pipeline struggles when other distractors stay still for long stretches (left).}
\label{fig:limit}
\vspace{-.6cm}
\end{figure}

%% file: sec/5_conclusion.tex
\section{Conclusion}
{
We present a robust pipeline for 3D reconstruction from casually captured 360$^\circ$ images using a consumer-grade dual-fisheye camera. We demonstrate that data collection for reconstruction can be performed more efficiently with dual-fisheye imagery than with commonly used perspective cameras due to their wider angular coverage. To leverage this advantage for scalable dataset collection, we address the central challenge of casual 360$^\circ$ capture—the always-visible capturer that violates photometric consistency—by reliably locating and masking the capturer for high-quality reconstruction. To benchmark this, we collect a dataset that provides a focused test-bed for robust 360$^\circ$ reconstruction.
}

While effective, our pipeline currently assumes fixed exposure and does not handle abrupt brightness changes or severe video motion blur; integrating recent advances that address these effects is a promising direction for future research~\cite{deutsch2026ppisp}. Further, our method struggles with \yf{static} distractors -- an uncommon scenario in typical casual dataset captures; see~\cref{fig:limit}. Nevertheless, FullCircle enables robust reconstruction across a wide range of casual captures, paving the way for scalable 360$^\circ$ reconstructable datasets that allow training feed-forward reconstruction networks from 360$^\circ$ imagery~\cite{360splatter, panosplatter}.

%% file: sec/X_suppl.tex
\clearpage
\setcounter{section}{0}
\renewcommand{\thesection}{\Alph{section}}
\renewcommand{\thesubsection}{\Alph{section}.\arabic{subsection}}
\section{Supplementary Materials}
Please refer to our website at \url{https://theialab.github.io/fullcircle} for additional \textbf{qualitative video results}.

We next report the corresponding quantitative results (LPIPS and SSIM) for our main experiments. We further present additional experiments that complement the main paper. Finally, we provide a detailed description of our masking pipeline, together with relevant implementation details, in~\cref{supp:details}.

\subsection{Additional image-quality metrics}
We complement our results in~\cref{sec:pers_comp} and~\cref{sec:fisheye_comp} with SSIM and LPIPS metrics. Specifically, we report \cref{supp:comparison-p_ssim,supp:comparison-p_lpips} as additional image-quality metrics for \cref{fig:comparison-p} in the main paper, and \cref{supp:comparison-f_ssim,supp:comparison-f_lpips} as additional metrics for \cref{fig:comparison-f}. 

\input{supp/comparison-p_ssim}

\input{supp/comparison-p_lpips}

\input{supp/comparison-f_ssim}

\input{supp/comparison-f_lpips}

\yf{\subsection{OmniSAM}
Here we provide additional qualitative results for OmniSAM~\cite{omnisam} to further illustrate that it is less robust at detecting the camera holder in highly distorted regions of omnidirectional images, often leaving residual operator pixels that appear as visible reconstruction artifacts, such as the reconstructed legs of the camera operator. In contrast, our masking approach applies SAMv2~\cite{sam} to synthetic fisheye frames centered on the camera operator, resulting in better segmentation and cleaner reconstructions.
\input{fig/omnisam_extra}}

\subsection{Using alternative tracking models (DINOv3~\cite{dinov3})}
\label{dino}
In the main paper, we obtain synthetic fisheye masks by automatically prompting SAMv2~\cite{sam} using the center of the first synthetic fisheye frame and propagating the mask over time. As an alternative, we also experiment with DINOv3~\cite{dinov3} for segmentation and tracking. However, DINOv3 requires an instance segmentation mask for the first frame, which prevents a fully automatic pipeline compared to our automated point-based SAMv2 prompting. For completeness, we report the DINOv3 results in \cref{supp:dino}, but do not adopt this variant as our main approach.
\input{supp/dino}

\subsection{Performance with unmasked calibration}
\label{unmasked-colmap}
For the main experiments, we provide COLMAP with both the fisheye frames and their corresponding masks. Without masking, we observed that in some scenes COLMAP extracts features on the capturer (e.g., in Room 3 or Persons), which leads to incorrect pose estimates once the person moves. To mitigate this, we apply the masks during calibration. For completeness, we also evaluate COLMAP with unmasked calibration; \cref{supp:colmap-not_masked} shows that masked calibration consistently improves the 3D reconstruction quality.
\input{supp/colmap-not_masked}

\yf{\subsection{Ricoh360 dataset}
Among existing datasets that contain the camera holder, we conduct experiments on the Ricoh360 dataset~\cite{egonerf} (Fig.~\ref{fig:egonerf}). We convert the omnidirectional inputs to dual-fisheye and apply our pipeline. The results show strong generalization: our method outperforms EgoNeRF and SLS by 1.3 dB and 0.9 dB in PSNR, respectively, across all scenes. Despite the challenging camera-holder scenario in Ricoh360, we reliably remove the capturer, producing clean reconstructions with empty regions where the geometry underneath the capturer was unobserved.

\input{fig/egonerf}

}

\subsection{Mask conversions}
\label{supp:details}
We provide a more complete description of our masking pipeline, including details on converting masks between perspective, fisheye, and omnidirectional camera models required by our method.

\paragraph{Sampling 16 pinhole cameras.}
For each omnidirectional image obtained from the camera software, we sample 16 virtual 90$^\circ$ pinhole cameras that together cover the full sphere. On each pinhole view, we run YOLOv8 followed by SAMv2 to detect and segment the person. From the resulting mask, we compute the center of mass and the mask area, since small false-positive masks can affect later steps. We then map these quantities back to the omnidirectional and the world space.

\paragraph{Reorienting the omni with the found direction.}
Once we have the per-pinhole centers of mass and their corresponding areas, we compute an area-weighted average of these centers in world coordinates to obtain a dominant direction for the capturer. We then re-orient the omnidirectional image so that this direction is centered.

\paragraph{Going to fisheye and back.}
We map the rotated 3D points $xyz$ to a synthetic fisheye frame, treating the rear fisheye as the reference and converting from the omnidirectional image to fisheye via

\begin{equation}
    \phi = \arcsin(-y), \qquad
    \lambda = \operatorname{atan2}(x, -z),
\end{equation}
\begin{equation}
    u = \left(1 + \frac{-\lambda}{\pi}\right)\, \frac{W}{2},
    \qquad
    v = \left(1 - \frac{2\,\phi}{\pi}\right)\, \frac{H}{2},
\end{equation}
where $\lambda$ and $\phi$ represent longitude and latitude, and $(u, v)$ represent omni image coordinates.
We then perform segmentation and tracking on these synthetic fisheye frames using SAMv2 (or DINOv3 in~\cref{dino}) to obtain fisheye masks. Finally, we map the masks back to the omnidirectional domain.

\paragraph{Going to original fisheye.}
For each fisheye camera, we recover its viewing directions and rotate them into world coordinates using the extrinsics obtained from a checkerboard calibration between the omnidirectional and fisheye frames. We then convert the resulting 3D points for the front and rear cameras, $\mathbf{x}_f$ and $\mathbf{x}_r$, into spherical coordinates $(\phi_f,\lambda_f)$ and $(\phi_r,\lambda_r)$. This yields the corresponding pixels in the omni image, which we map back to the original fisheye frames as:

\begin{align*}
    u_f &= \left(1 + \frac{-\lambda_f + \pi}{\pi}\right)\frac{W}{2}, &
    v_f &= \left(1 - \frac{2\phi_f}{\pi}\right)\frac{H}{2}, \\
    u_r &= \left(1 + \frac{-\lambda_r}{\pi}\right)\frac{W}{2}, &
    v_r &= \left(1 - \frac{2\phi_r}{\pi}\right)\frac{H}{2}.
\end{align*}

%% file: supp/comparison-p_ssim.tex
\begin{table*}[!ht]
\centering
\setlength{\tabcolsep}{8pt}
\resizebox{\textwidth}{!}{
\begin{tabular}{lc|ccccccccc|c}
\textbf{Method} & Robust & Room 1 & Flat 1 & Flat 2 & Room 2 & Room 3 & Lab & Lounge & Persons & Dark & \textbf{Mean} \\
\midrule
3DGS~\cite{3dgs}              & \xmark & 0.90 & 0.83 & 0.85 & 0.90 & 0.85 & 0.80 & 0.72 & 0.82 & 0.87 & 0.84\\
SLS-MLP~\cite{spotlesssplats} & \cmark & 0.90 & 0.83 & 0.86 & 0.91 & 0.86 & 0.80 & 0.71 & 0.83 & 0.89 & 0.84\\
NOTG~\cite{nerf_on_the_go}    & \cmark & 0.80 & 0.69 & 0.86 & 0.80 & 0.80 & 0.70 & 0.57 & 0.84 & 0.84 & 0.77\\
\midrule
\textbf{FullCircle (Ours)}    & \cmark & \textbf{0.92} & \textbf{0.86} & \textbf{0.89} & \textbf{0.93} & \textbf{0.88} & \textbf{0.83} & \textbf{0.75} & \textbf{0.86} & \textbf{0.90} & \textbf{0.87}\\
\end{tabular}
}
\caption{
\textbf{Perspective} -- SSIM~$\uparrow$ metric for reconstruction quality using undistorted fisheye images vs. ours.
}
\vspace{-1.2cm}
\label{supp:comparison-p_ssim}
\end{table*}

%% file: supp/comparison-p_lpips.tex
\begin{table*}[!ht]
\centering
\setlength{\tabcolsep}{8pt}
\resizebox{\textwidth}{!}{
\begin{tabular}{lc|ccccccccc|c}
\textbf{Method} & Robust & Room 1 & Flat 1 & Flat 2 & Room 2 & Room 3 & Lab & Lounge & Persons & Dark & \textbf{Mean} \\
\midrule
3DGS~\cite{3dgs}              & \xmark & 0.22 & 0.35 & 0.26 & 0.25 & 0.31 & 0.30 & 0.26 & 0.37 & 0.30 & 0.29\\
SLS-MLP~\cite{spotlesssplats} & \cmark & \textbf{0.19} & 0.31 & 0.22 & 0.23 & 0.28 & 0.30 & 0.28 & 0.33 & \textbf{0.25} & 0.27\\
NOTG~\cite{nerf_on_the_go}    & \cmark & 0.31 & 0.50 & 0.23 & 0.41 & 0.34 & 0.41 & 0.45 & \textbf{0.30} & 0.28 & 0.36\\
\midrule
\textbf{FullCircle (Ours)}    & \cmark & \textbf{0.19} & \textbf{0.30} & \textbf{0.18} & \textbf{0.21} & \textbf{0.26} & \textbf{0.27} & \textbf{0.23} & 0.31 & \textbf{0.25} & \textbf{0.25}\\
\end{tabular}
}
\caption{
\textbf{Perspective} -- LPIPS~$\downarrow$ metric for reconstruction quality using undistorted fisheye images vs. ours.
}
\vspace{-1.2cm}
\label{supp:comparison-p_lpips}
\end{table*}

%% file: supp/comparison-f_ssim.tex
\begin{table*}[!ht]
\centering
\setlength{\tabcolsep}{8pt}
\resizebox{\textwidth}{!}{
\begin{tabular}{lc|ccccccccc|c}
\textbf{Method} & Robust & Room 1 & Flat 1 & Flat 2 & Room 2 & Room 3 & Lab & Lounge & Persons & Dark & \textbf{Mean} \\
\midrule
3DGRT~\cite{3dgrt}            & \xmark & 0.94 & 0.90 & 0.91   & 0.93 & 0.91 & 0.89          & 0.85          & 0.89 & 0.91                 & 0.90\\
SLS-MLP~\cite{spotlesssplats} & \cmark & 0.94 & \textbf{0.91} & 0.93 & 0.94 & \textbf{0.92} & \textbf{0.90} & 0.85 & 0.90 & \textbf{0.92} & 0.91\\
OLRF~\cite{omnilocalrf}       & \cmark & 0.82 & 0.74 & 0.81   & 0.82 & 0.78 & 0.68          & 0.63          & 0.75 & 0.81                 & 0.76\\
NOTG~\cite{nerf_on_the_go}    & \cmark & 0.86 & 0.73 & 0.83   & 0.92 & 0.86 & 0.76          & 0.69          & 0.84 & 0.81                 & 0.81\\
\midrule
\textbf{FullCircle (Ours)}    & \cmark & \textbf{0.95} & \textbf{0.91} & \textbf{0.94} & \textbf{0.95} & \textbf{0.92} & \textbf{0.90} & \textbf{0.87} & \textbf{0.91} & \textbf{0.92} & \textbf{0.92}\\
\end{tabular}
}
\caption{
\textbf{Fisheye} -- SSIM~$\uparrow$ metric for reconstruction quality using fisheye images. \yf{For OmniLocalRF, we report results using omnidirectional images.}
}
\vspace{-1.2cm}
\label{supp:comparison-f_ssim}
\end{table*}

%% file: supp/comparison-f_lpips.tex
\begin{table*}[!ht]
\centering
\setlength{\tabcolsep}{8pt}
\resizebox{\textwidth}{!}{
\begin{tabular}{lc|ccccccccc|c}
\textbf{Method} & Robust & Room 1 & Flat 1 & Flat 2 & Room 2 & Room 3 & Lab & Lounge & Persons & Dark & \textbf{Mean} \\
\midrule
3DGRT~\cite{3dgrt}            & \xmark & 0.10 & 0.19 & 0.12 & 0.14          & 0.16          & 0.16          & 0.12 & 0.20 & 0.16          & 0.15\\
SLS-MLP~\cite{spotlesssplats} & \cmark & 0.09 & 0.17 & 0.10 & \textbf{0.11} & \textbf{0.15} & \textbf{0.14} & 0.12 & 0.19 & \textbf{0.14} & 0.14\\
OLRF~\cite{omnilocalrf}       & \cmark & 0.40 & 0.52 & 0.42 & 0.41          & 0.49          & 0.50          & 0.48 & 0.47 & 0.46          & 0.46\\
NOTG~\cite{nerf_on_the_go}    & \cmark & 0.19 & 0.47 & 0.22 & 0.14          & 0.23          & 0.30          & 0.35 & 0.25 & 0.30          & 0.27\\
\midrule
\textbf{FullCircle (Ours)}    & \cmark & \textbf{0.08} & \textbf{0.16} & \textbf{0.09} & \textbf{0.11} & \textbf{0.15} & \textbf{0.14} & \textbf{0.10} & \textbf{0.18} & \textbf{0.14} & \textbf{0.13}\\
\end{tabular}
}
\caption{
\textbf{Fisheye} -- LPIPS~$\downarrow$ metric for reconstruction quality using fisheye images.
\yf{For OmniLocalRF, we report results using omnidirectional images.}
}
\vspace{-1.2cm}
\label{supp:comparison-f_lpips}
\end{table*}

%% file: fig/omnisam_extra.tex
\begin{figure}[!ht]
\centering
\includegraphics[width=\textwidth]{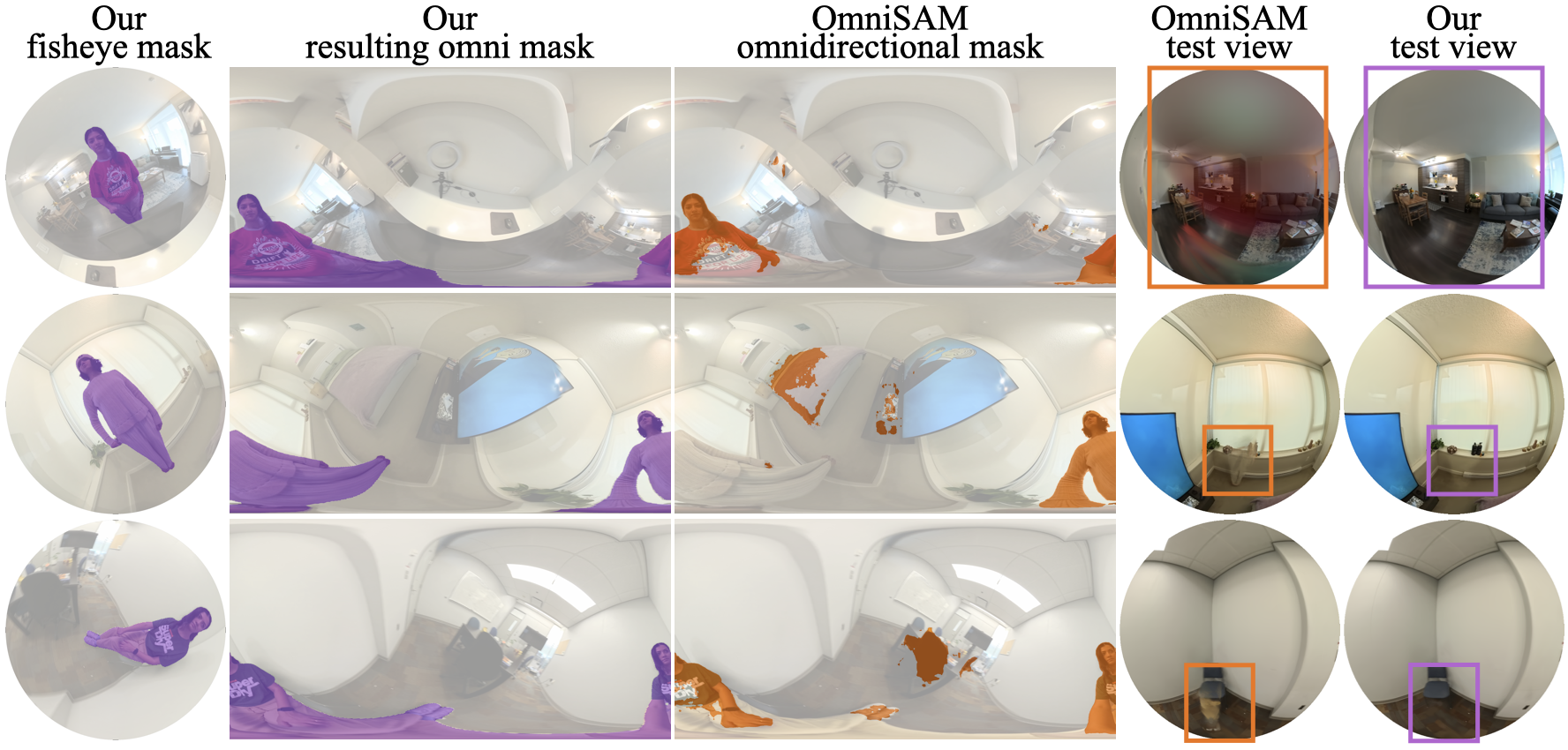}
\captionof{figure}
{\textbf{Camera operator segmentation} -- 
Additional qualitative comparison between OmniSAM and our method for camera operator segmentation.}
\vspace{-1cm}
\label{fig:omnisam_extra}
\end{figure}

%% file: supp/dino.tex
\begin{table*}[!ht]
\centering
\setlength{\tabcolsep}{8pt}
\resizebox{\textwidth}{!}{
\begin{tabular}{l|ccccccccc|c}
\textbf{Method} & Room 1 & Flat 1 & Flat 2 & Room 2 & Room 3 & Lab & Lounge & Persons & Dark & \textbf{Mean} \\
\midrule
DINOv3~\cite{dinov3} & 32.59 & \textbf{29.21} & 30.50 & 31.61 & 29.87 & 29.43 & 24.35 & 30.86 & \textbf{27.99} & 29.60\\

\midrule
\textbf{FullCircle (Ours)} & \textbf{32.64} & 29.11 & \textbf{30.55} & \textbf{31.63} & \textbf{29.91} & \textbf{29.54} & \textbf{24.40} & \textbf{30.99} & 27.79 & \textbf{29.62}\\
\end{tabular}
}
\caption{
\textbf{DINOv3} -- Reconstruction quality reported in PSNR~$\uparrow$ metric for DINOv3 masking alternative to SAMv2.
}
\label{supp:dino}
\end{table*}

%% file: supp/colmap-not_masked.tex
\begin{table*}[!ht]
\centering
\setlength{\tabcolsep}{8pt}
\resizebox{\textwidth}{!}{
\begin{tabular}{lc|ccccccccc|c}
\textbf{Method} & COLMAP & Room 1 & Flat 1 & Flat 2 & Room 2 & Room 3 & Lab & Lounge & Persons & Dark & \textbf{Mean} \\
\midrule
3DGRT~\cite{3dgrt} & Unmasked & 27.56 & 25.75 & 24.99 & 27.88 & 27.48 & 26.90 & 23.43 & failed & 25.51 & -\\
FullCircle (Ours)  & Unmasked & 32.20 & 28.31 & 30.20 & 31.30 & 28.80 & 28.68 & 24.38 & failed & 27.77 & -\\

\midrule
3DGRT~\cite{3dgrt}         & Masked &  27.37 & 26.81 & 25.09 & 27.92 & 28.41 & 26.66 & 23.41 & 27.56 & 25.71 & 26.55\\
\textbf{FullCircle (Ours)} & Masked & \textbf{32.64} & \textbf{29.11} & \textbf{30.55} & \textbf{31.63} & \textbf{29.91} & \textbf{29.54} & \textbf{24.40} & \textbf{30.99} & \textbf{27.79} & \textbf{29.62}\\
\end{tabular}
}
\caption{
\textbf{Unmasked calibration} -- Reconstruction performance reported in PSNR~$\uparrow$ metric using unmasked calibration.
}
\vspace{-1cm}
\label{supp:colmap-not_masked}
\end{table*}

%% file: fig/egonerf.tex
\begin{figure*}[!ht]
\centering
\includegraphics[width=\textwidth]{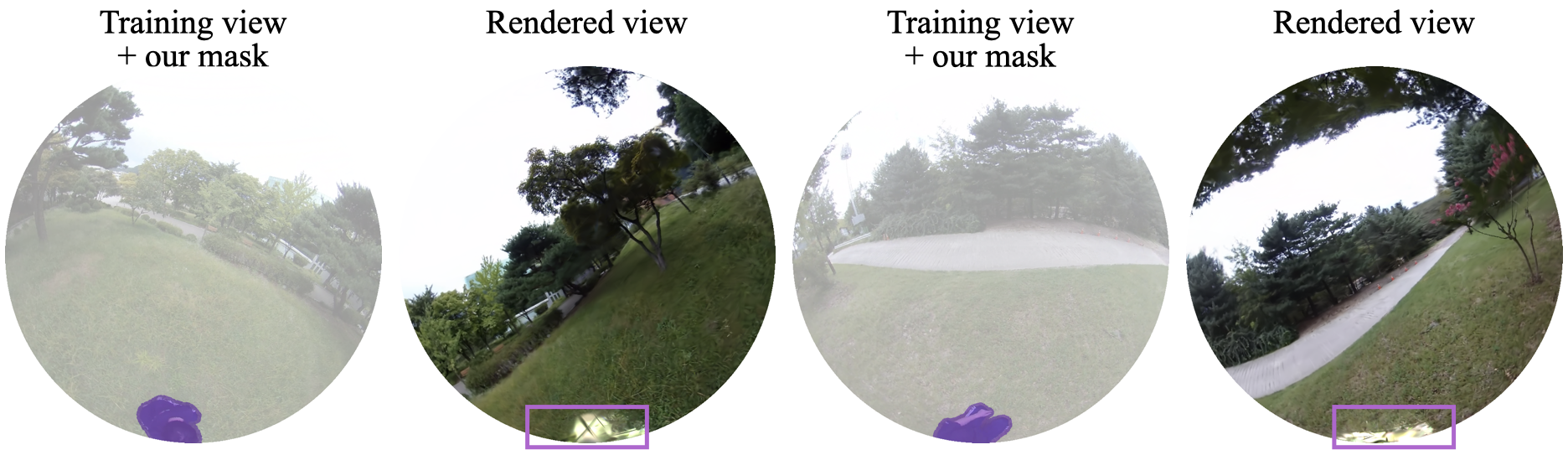}
\setlength{\tabcolsep}{8pt}
\resizebox{\textwidth}{!}{
\begin{tabular}{l|ccccccccccc|c}
\textbf{Method} & Bricks & Bridge & BridgeUnder & CatTower & Center & Farm & Flower & GalleryChair & GalleryPillar & Garden & Poster & \textbf{Mean}\\
\midrule
EgoNerf-100k~\cite{egonerf} (from paper) & 23.37 & 23.40 & 24.94 & 24.23 & 28.45 & 22.23 & 21.80 & 27.78 & 28.02 & 26.87 & 26.62 & 25.25\\
SLS-MLP~\cite{spotlesssplats} & 25.08 & 22.60 & 25.61 & 25.66 & 27.30 & 21.02 & 23.12 & 28.60 & 28.06 & 26.65 & 27.73 & 25.58\\
\midrule
\textbf{FullCircle (Ours)} &\textbf{25.93}&\textbf{24.70}&\textbf{26.15}&\textbf{26.26}&\textbf{28.67}&\textbf{22.86}&\textbf{23.96}&\textbf{29.13}&\textbf{28.41}&\textbf{27.28}&\textbf{28.09}&\textbf{26.50}\\
\end{tabular}
}
\caption{
\textbf{Ricoh360 dataset} -- Quantitative comparison on the Ricoh360 dataset~\cite{egonerf} together with qualitative results of our method. Our approach achieves higher PSNR~$\uparrow$ in all scenes and reliably removes the static camera holder. Regions underneath the holder remain empty in the reconstruction, since they are unobserved in the input, rather than producing visible artifacts.
}
\label{fig:egonerf}
\end{figure*}